\documentclass[preprint]{elsarticle}
\usepackage{graphicx} 
\usepackage{hyperref}
\usepackage{multirow}
\usepackage{color}
\usepackage{amsmath}
\usepackage{combelow}

\definecolor{carminepink}{rgb}{0.92, 0.3, 0.26}
\definecolor{cobalt}{rgb}{0.0, 0.28, 0.67}
\definecolor{cerulean}{rgb}{0.0, 0.48, 0.65}

\newcount\Comments
\newcommand{\kibitz}[2]{\ifnum\Comments=1\textcolor{#1}{#2}\fi}

\journal{ArXiV}

\begin{document}

\begin{frontmatter}

\title{Convolution Operator Network for Forward \\ and Inverse Problems (FI-Conv):\\ Application to Plasma Turbulence Simulations}

\author[inst1]{Xingzhuo Chen}

\affiliation[inst1]{organization={Texas A\&M Institute of Data Science},
            city={College Station},
            postcode={77843}, 
            state={TX},
            country={USA}}

\author[inst2]{Anthony Poole}

\affiliation[inst2]{organization={Department of Physics, The University of Texas at Austin},
            city={Austin},
            postcode={78705}, 
            state={TX},
            country={USA}}
            
\author[inst3]{Ionu\cb{t}-Gabriel Farca\cb{s}}
\affiliation[inst3]{%
  organization={Department of Mathematics and Division of Computational Modeling and Data Analytics, Academy of Data Science, Virginia Tech},
  city={Blacksburg},
  state={VA},
  postcode={24061},
  country={USA}
}

\author[inst4]{David R. Hatch}
\affiliation[inst4]{%
  organization={Institute for Fusion Studies, The University of Texas at Austin},
  city={Austin},
  state={TX},
  postcode={78705},
  country={USA}
}

\author[inst1,inst5]{Ulisses Braga-Neto}
\affiliation[inst5]{organization={Department of Electrical and Computer Engineering, Texas A\&M University},
            city={College Station},
            postcode={77843}, 
            state={TX},
            country={USA}}

\begin{abstract}
    We propose the Convolutional Operator Network for Forward and Inverse Problems (FI-Conv), a framework capable of predicting system evolution and estimating parameters in complex spatio-temporal dynamics, such as turbulence.
    FI-Conv is built on a U-Net architecture, in which most convolutional layers are replaced by ConvNeXt V2 blocks. 
    This design preserves U-Net’s performance on inputs with high-frequency variations while maintaining low computational complexity.
    FI-Conv uses an initial state, PDE parameters, and evolution time as input to predict the system’s future state. 
    As a representative example of a system exhibiting complex dynamics, we evaluate the performance of FI-Conv on the task of predicting turbulent plasma fields governed by the Hasegawa–Wakatani (HW) equations.
    The HW system models two-dimensional electrostatic drift-wave turbulence and exhibits strongly nonlinear behavior, making accurate approximation and long-term prediction particularly challenging.
    Using an autoregressive forecasting procedure, FI-Conv achieves accurate forward prediction of the plasma state evolution over short times ($t\sim 3$) and captures the statistical properties of derived physical quantities of interest over longer times ($t\sim 100$). 
    Moreover, we develop a gradient-descent–based inverse estimation method that accurately infers PDE parameters from plasma state evolution data, without modifying the trained model weights. 
    Collectively, our results demonstrate that FI-Conv can be an effective alternative to existing physics-informed machine learning methods for systems with complex spatio-temporal dynamics.
\end{abstract}






\begin{keyword}
  Operator Learning \sep Hasegawa-Wakatani Equations \sep Plasma Turbulence \sep
  Parameter Estimation
\end{keyword}

\end{frontmatter}

\section{Introduction}
\noindent
Physics-informed neural networks (PINNs) are an emerging scientific machine learning approach for solving nonlinear partial differential equations (PDEs).~\cite{DPT1994,lagaris1998artificial,Raissi2019PINN}.
Although PINNs have been successfully applied to a variety of PDEs, including the radiative transfer equation~\cite{Mishra2021PINNRT}, the Burgers equation~\cite{Zhang2020Burgers}, and the Navier–Stokes equations~\cite{Dwivedi2019Navier}, their training and inference are typically restricted to a single PDE instance with fully specified boundary conditions and physical parameters.
As a result, extending PINNs to multiple problem instances can become computationally very expensive.
To overcome these limitations, neural operators~\cite{Lu2019DON,Li2020FNO} were introduced to learn mappings between function spaces, enabling the direct approximation of solution operators for parametric PDEs. 
For example, Fourier Neural Operators (FNOs)~\cite{Li2020FNO} leverage spectral representations to improve training efficiency and enable resolution-invariant prediction of PDE solution fields.
Once trained, a neural operator can in principle be applied to multiple instances of a PDE across varying parameters or inputs without requiring further updates to the model weights. 
Moreover, neural operators may be trained purely from data or augmented with physics-informed loss terms, in the fashion of PINNs~\cite{Xu2022TLPIDON,Li2021PINO,Goswami2022PINO}.


While many machine learning approaches are primarily developed for forward PDE modeling, scientific and engineering applications oftentimes require solving inverse problems, in which hidden or latent parameters must be inferred from partial or indirect observations~\cite{Chen2020AIAI}.
In contrast to the forward problem, where the goal is to numerically solve a well-posed PDE given complete physical specifications, the inverse problem is typically more challenging. 
Solutions usually fail to satisfy Hadamard’s criteria of well-posedness, namely existence, uniqueness, and continuous dependence on the data, rendering the inverse problem ill-posed and necessitating additional regularization~\cite{tarantola2005inverse,isakov2018inverse}.
To address the challenges of solving inverse problems with neural networks, diffusion-based neural networks have been proposed to reconstruct physical fields from sparsely sampled experimental data by enforcing physics-informed loss functions~\cite{Huang2024DiffuPDE,Kacmaz2025Diffu,Utkarsh2025Diffu}.
Graph neural networks have also been applied to solving inverse problems, with demonstrated success on benchmark systems such as the Navier–Stokes equations~\cite{Kumar2023GNNInverse}.
In addition, Bayesian formulations combined with neural operators have been developed to solve inverse problems while providing uncertainty quantification for PDE parameters~\cite{Yang2021BPINN,Li2023BPINN,Raj2025Inverse}.
Nevertheless, as highlighted in a recent survey on operator learning and inverse problems~\cite{nelsen2025operatorlearningmeetsinverse}, inverse problem solutions based on neural operators can remain computationally challenging.

To address the computational challenges associated with both forward prediction of parametric PDE systems and inverse parameter estimation, we propose the \emph{Convolution Operator Network for Forward and Inverse Problems (FI-Conv)}. 
FI-Conv is designed to handle systems with complex spatio-temporal dynamics, such as turbulence.
Key input parameters (i.e., initial state, PDE parameters, and evolution time) are embedded directly into the network architecture, allowing a single trained FI-Conv model to perform forward prediction and parameter estimation across a range of parameter values without requiring additional weight updates.
Architecturally, FI-Conv combines a U-Net backbone~\cite{Ronneberger2015Unet} with ConvNeXt V2 blocks~\cite{Woo2023Convnext}, and is deployed in an autoregressive setting to achieve long-term prediction of the PDE state evolution.

As a representative example of a complex application, we consider a modified version of the Hasegawa–Wakatani (HW) equation system~\cite{Hasegawa1983HW}.
The HW equations model two-dimensional drift-wave turbulence in collision-dominated plasmas within a slab geometry. 
Despite its simplified setting, the HW model captures key physical mechanisms underlying plasma turbulence and linear instabilities, and has been widely used to study turbulent transport near the edge of tokamak devices under strong magnetic fields (e.g., Ref.~\cite{Guillon2025PhaseTransition,Lin2025Tessellation}).
Owing to its nonlinear and multiscale dynamics, the HW system provides a challenging benchmark for evaluating the performance of machine learning--based methods.

Forward modeling of plasma field evolution in the HW system has been explored using a range of machine learning approaches, including data-driven reduced-order modeling~\cite{Gahr2024HWROM}, FNOs~\cite{Gopakumar2024FNOPlasma,Carey2025FNOPlasma,Pamela2025FNOPlasma}, autoencoder-based architectures~\cite{Clavier2025VAEHW,Khrabry2025HEAP}, and recurrent neural networks~\cite{Artigues2025HW2D}.
In particular, Gahr et al.~\cite{Gahr2024HWROM} developed scientific machine-learning–based reduced models to predict the time evolution of density and electrostatic potential fields, as well as derived quantities of interest, and examined model performance across multiple instances of the adiabatic parameter $c_1$. 
Clavier et al.~\cite{Clavier2025VAEHW} employed variational autoencoder–based generative models to construct surrogate representations of the electrostatic potential field.
Moreover, Artigues et al.~\cite{Artigues2025HW2D} presented one of the first machine learning studies to assess predictive generalization of HW dynamics to previously unseen values of the adiabatic coefficient $c_1$.

Unlike existing approaches that primarily focus on forward prediction with variations in a single input parameter, the proposed FI-Conv framework handles simultaneous variations in four PDE parameters, enabling efficient forward prediction of both the system state and derived quantities. 
To our knowledge, inverse parameter estimation has not previously been addressed for the HW system.
FI-Conv addresses this gap. 
Together, these capabilities can facilitate calibrated reduced-order models and a range of many-query scientific and engineering applications, such as uncertainty quantification and parameter studies.

This paper is organized as follows. 
Section~\ref{sec:neuralnet} presents the FI-Conv neural network architecture.
Section~\ref{sec:HW_equations} describes the HW equations and the numerical procedures used to generate training and testing data. 
Section~\ref{sec:hyperparameter} details the FI-Conv hyperparameters and training strategy for the HW problem. 
Section~\ref{sec:forward} presents numerical experiments on forward prediction of plasma state evolution using the trained FI-Conv model. 
In Sec.~\ref{sec:inverse}, FI-Conv is applied to inverse parameter estimation for the HW system using sample data, without updating the model weights. 
Finally, Sec.~\ref{sec:conclusion} provides concluding remarks and outlines directions for future work.
The code used for all simulations is publicly available at \href{https://github.com/GeronimoChen/PIHW}{https://github.com/GeronimoChen/PIHW}.

\section{The proposed FI-Conv architecture}\label{sec:neuralnet}

The proposed Convolutional Operator Network for Forward and Inverse Problems (FI-Conv) is based on a U-Net architecture~\cite{Ronneberger2015Unet}. 
In this design, the standard convolutional layers in the encoder blocks are replaced with ConvNeXt~V2 blocks~\cite{Woo2023Convnext}, while the input convolution layer and the transposed convolution layers in the decoder blocks remain unchanged.
The skip connections inherited from U-Net enable FI-Conv to accurately capture small-scale spatial variations in the data. 
Compared to conventional convolutional layers, ConvNeXt V2 blocks require fewer trainable parameters and lower computational cost, while achieving improved performance in computer vision tasks~\cite{Woo2023Convnext}. 

\begin{figure}
    \centering
    \includegraphics[width=\linewidth]{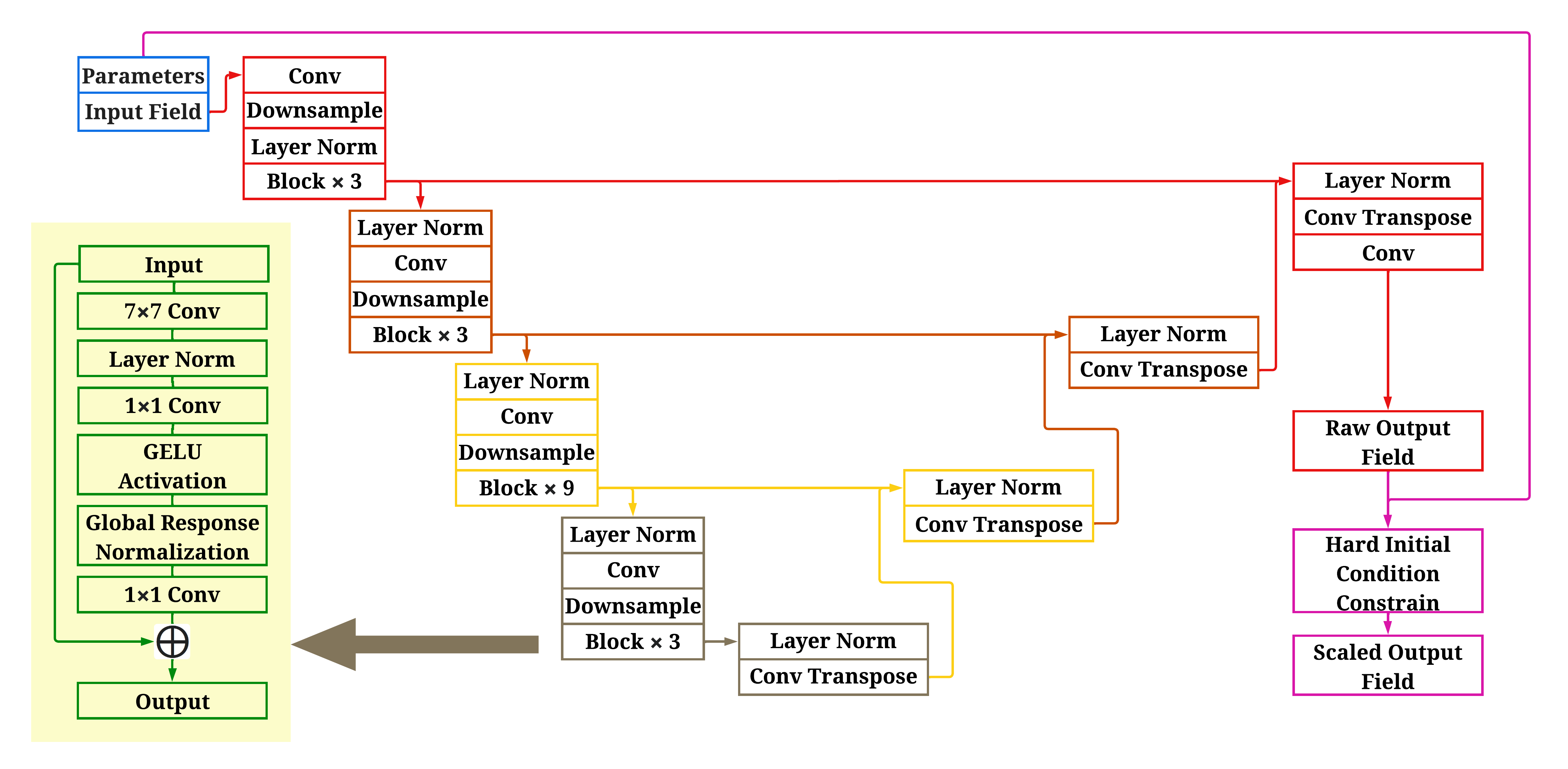}
    \caption{Schematic of the proposed FI-Conv architecture. 
       ConvNeXt~V2 blocks are shown in green (left). The input field and embedded parameters, including the evolution time and PDE parameters, are shown in blue.
       The first, second, third, and fourth depth levels of the U-Net encoder–decoder hierarchy are indicated in red, brown, yellow, and gray, respectively. 
       The enforcement of hard initial-condition constraints is highlighted in purple. }
    \label{fig:FI-Conv-Detail}
\end{figure}

Figure~\ref{fig:FI-Conv-Detail} illustrates the overall FI-Conv architecture.
As shown in blue, the input to FI-Conv consists of two components: (1) an input field, representing the two-dimensional initial state of the PDE variables, and (2) parameters, including the PDE parameters and the evolution time $t$.
The evolution time is ``bottleneck-injected" at the input layer together with hard initial-condition constraints~\cite{Hao2024HardPINN}. 
This design enables FI-Conv to accurately predict the system state at time $t$
for short integration horizons, and to efficiently generate long-term predictions through iterative autoregressive rollout.
The PDE parameters are similarly injected at the input layer, allowing inference under varying physical conditions. 
This formulation also enables automatic differentiation of the network output with respect to the PDE parameters, thereby facilitating solving parameter inference problems.

Consistent with many U-Net--based architectures used in computer vision (e.g., Refs.~\cite{Ronneberger2015Unet,Ovadia2025tcunet}), the FI-Conv encoder comprises four $2\times2$ downsampling layers, while the decoder consists of four transposed convolution layers that restore the spatial resolution at each hierarchical level.
For the input resolution of $128\times128$ used in our numerical experiments (see Sec.~\ref{sec:HW_equations}), this configuration yields a coarsest latent representation of size $8\times8$.
At this level, the $7\times7$ convolution kernels in the ConvNeXt~V2 blocks effectively access global spatial information.
The padding strategy for all convolutional, ConvNeXt~V2, and transposed convolution layers is selected to match the boundary conditions of the underlying PDE system.
Specifically, zero padding is used for Dirichlet or Neumann boundary conditions, circular padding for periodic boundary conditions, and reflective padding for symmetric boundary conditions.
FI-Conv further incorporates a hard boundary-condition enforcement strategy~\cite{Hao2024HardPINN}, which ensures that initial conditions are exactly satisfied and improves prediction accuracy. 
This component is highlighted in magenta in Fig.~\ref{fig:FI-Conv-Detail}.

\begin{figure}
    \centering
    \includegraphics[width=\linewidth]{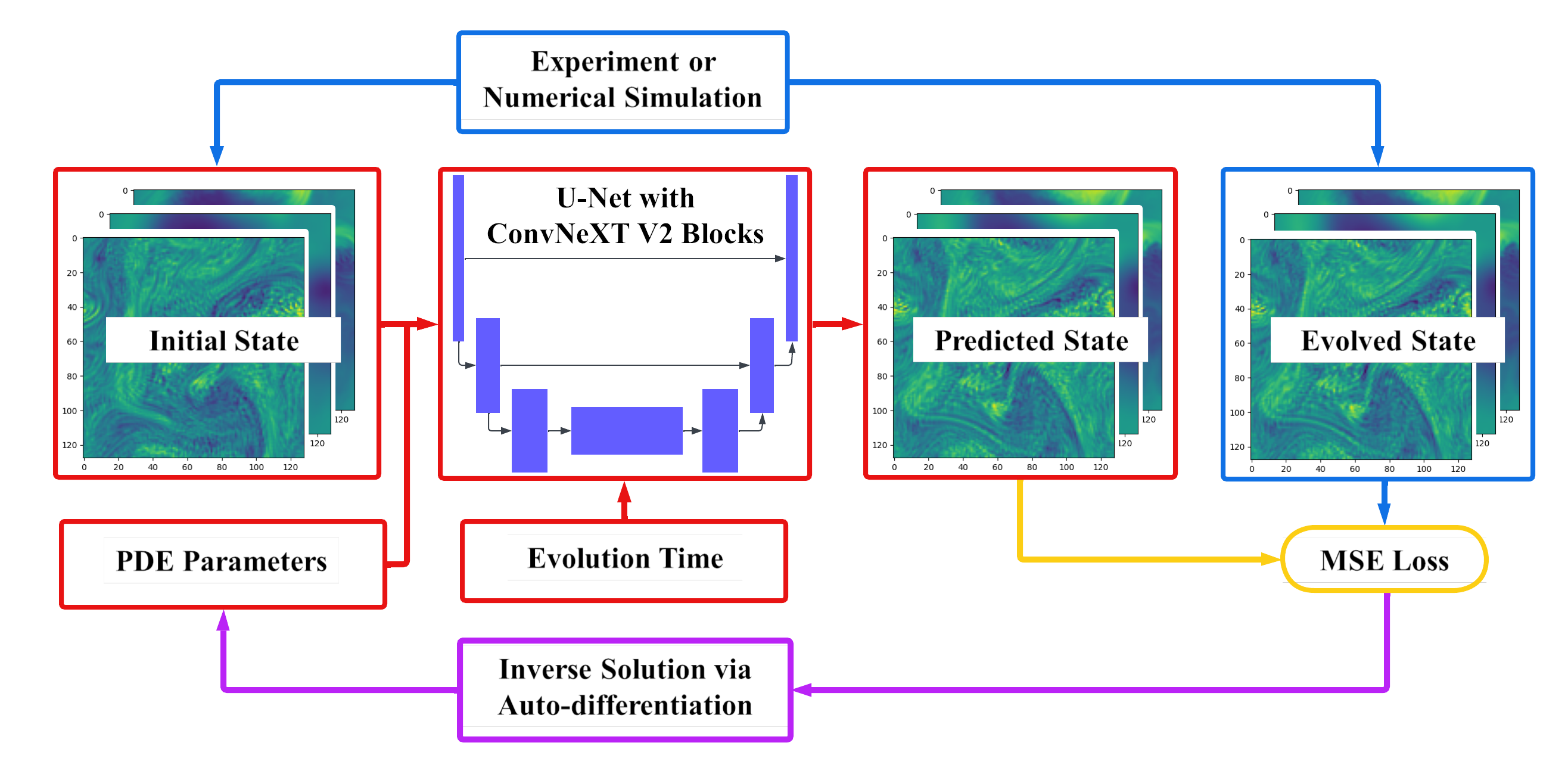}
    \caption{Overview of the FI-Conv workflow. Blue arrows denote experimental or numerical simulation data, red arrows indicate FI-Conv forward prediction, yellow arrows represent the training process, and purple arrows correspond to the inverse parameter estimation problem. 
    }
    \label{fig:FI-Conv}
\end{figure}

Figure~\ref{fig:FI-Conv} provides a high-level overview of the FI-Conv workflow.
First, training data are generated from experiments or numerical simulations (blue).
Then, FI-Conv performs forward prediction, mapping an initial state of the PDE variables to an evolved state under specified PDE parameters and evolution time (red).
Third, the predicted evolved state is compared with the corresponding numerical solution, and FI-Conv is trained using a mean-squared error (MSE) loss function (yellow).
Finally, the trained FI-Conv model is applied to inverse problems, namely the estimation of PDE parameters from new data by minimizing the MSE loss through automatic differentiation and gradient-based optimization, without necessitating updating the network weights (magenta).
The network input consists of an initial state, PDE parameters, and evolution time. 
Treating the PDE parameters as inputs enables efficient inverse modeling by allowing parameter estimation directly from input–output pairs via automatic differentiation.

FI-Conv can be used both in a purely data-driven regime, as adopted in the present work, and in a physics-informed setting.
When the evolution time is included as an input, temporal derivatives of the network output can be computed via automatic differentiation.
Spatial derivatives can be approximated using finite-difference operators applied to neighboring grid points in the two-dimensional output field.
This design enables the incorporation of physics-based loss terms without requiring changes to the underlying network architecture.


\section{The Hasegawa-Wakatani model for resistive drift-wave turbulence} \label{sec:HW_equations}
In this work, we evaluate FI-Conv using the HW model for predicting drift-wave plasma turbulence. Due to its intrinsically nonlinear and turbulent dynamics, the HW system provides a representative testbed for assessing FI-Conv’s performance in both forward simulation and inverse problem settings. 
We focus on this model because its self-sustained turbulence, driven by drift instabilities from finite plasma resistivity and background density gradients, presents significant challenges for machine-learning–based approximations.
Section~\ref{subsec:HW_equations_and_QoI} presents the HW system of equations and the derived quantities of interest. 
Section~\ref{subsec:HW_training} describes the generation of the training and testing datasets.

\subsection{The Hasegawa-Wakatani equations and derived quantities of interest} \label{subsec:HW_equations_and_QoI}

The HW equation system describes the time evolution of the fluctuations of the plasma electrostatic potential $\phi$, density $n$, and vorticity $\Omega$, all of which are functions of 2-D normalized coordinates (x, y) and time (t)~\cite{camargoResistiveDriftWave1995,Majda2018SpaFour}: 
\begin{eqnarray}
    \frac{\partial \Omega}{\partial t} &=& c_1 (\phi-n) -c_{pb}\left[\phi , \Omega\right] +\nu \nabla^{2N}n \\  
    \frac{\partial n}{\partial t} &=& c_1 (\phi-n) - c_{pb}\left[\phi , n\right] - \kappa \frac{\partial \phi}{\partial y} + \nu \nabla^{2N}\Omega \, \\
    \Omega &=& \nabla^2\phi \ ,
\end{eqnarray}
where the square bracket represents the Poisson operator 
\begin{equation}
    [p,q]=\frac{\partial p}{\partial x}\frac{\partial q}{\partial y}-\frac{\partial p}{\partial y}\frac{\partial q}{\partial x} \,, 
\end{equation}
$c_1$ denotes the adiabaticity coefficient that changes the dynamics from hydrodynamic limit ($c_1\rightarrow0$) to the adiabatic limit ($c_1\rightarrow+\infty$), and $c_{pb}$ controls the strength of the Poisson bracket.
Moreover, $\kappa = -\partial/\partial_x \mathrm{ln} (n_0)$ is the inverse density gradient scale length, where $n_0$ denotes the background density.  
The hyperdiffusion terms $\nabla^{2N}$ are included to suppress the accumulation of energy at the grid scale, with the hyperdiffusion coefficient $\nu$ controlling the strength of small-scale dissipation.
The boundary conditions are periodic in both $x$ and $y$ directions.  
The following normalizations are employed: $x \rightarrow x/\rho_s$, $y \rightarrow y/\rho_s$, $t \rightarrow t\Omega_i$, $\phi \rightarrow e \phi/T_0$, and $n \rightarrow n/n_0$, where $\rho_s = c_s/ \Omega_i$ represents the gyroradius, $c_s = \sqrt{T_0/m_i}$ is the sound speed, $T_0$ denotes the background temperature, $\Omega_i = eB/m_i$ is the ion cyclotron frequency, and $m_i$ denotes the ion mass. 

The adiabaticity parameter is defined as $c_1 = \frac{T_0}{n_0 \eta_{||} e^2}\frac{k_{||}^2}{\Omega_i}$, where $k_{||}$ represents the parallel (to the magnetic field) wavenumber and $\eta_{||}$ the parallel conductivity. 
We note that the parameter $c_{pb}$ is not a physical parameter of the HW equations; rather, we introduce it here as a means to explore additional variations in the dynamics of this system and thereby assess the ability of FI-Conv to generalize across a broader range of dynamical scenarios. 
This parameter effectively modulates the relative strength of the nonlinear terms. 
The standard HW equations are recovered when $c_{pb} = 1$, which corresponds to a subset of the simulations considered in the present paper.


After initialization, the dynamics undergo a transient phase in which linear drift-waves dominate, and the energy grows exponentially. 
The dynamics subsequently enter the turbulent phase, where the effect of quadratic nonlinearities becomes significant.

Physically relevant quantities of interest include $\Gamma_n$ and $\Gamma_c$, calculated as: 
\begin{equation}
\begin{aligned}
    \Gamma_n &\,=\, -\int_x\int_y dx\ dy\ n(x,y)\frac{\partial \phi(x,y)}{\partial y}\,,  \\[0.5ex] 
    \Gamma_c &\,=\, c_1 \int_x\int_y dx\ dy\ \left(n(x,y)-\phi(x,y)\right)^2 \,. 
\end{aligned}
\end{equation}
$\Gamma_n$ measures the radial particle flux---$E \times B$ advection of the density fluctuations. 
Energy is simultaneously dissipated through resistive effects, characterized by the resistive dissipation rate $\Gamma_c$. 

\subsection{Generating the training and testing datasets} \label{subsec:HW_training}
We employ the numerical solver HW2D~\cite{Greif2023JOSSHW2D} to generate the training and testing datasets for FI-Conv. 
The simulation domain size $L$ is determined by the parameter $k_0$ via $L = 2\pi/k_0$. 
The initial conditions are realizations of 2D Gaussian random fields.
The Poisson brackets are discretized using a fourth--order Arakawa scheme~\cite{arakawaComputationalDesignLongTerm1997}. 
The remaining spatial derivatives are discretized using second--order finite differences.
Time integration is performed using a fourth-order explicit Runge--Kutta scheme, followed by solving the Poisson equation, $\nabla^2 \phi = f$, via spectral methods to obtain the potential field $\phi$.
The hyperdiffusion parameters are set to $\nu=5\times 10^{-10}$ and $N=3$ for numerical stability.

\begin{figure}
    \centering
    \includegraphics[width=0.24\linewidth]{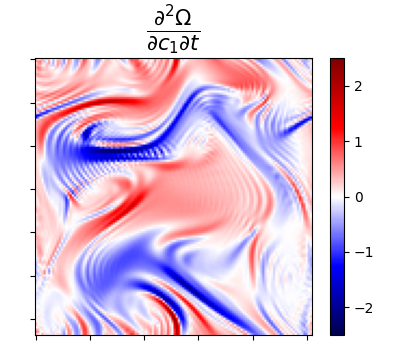}
    \includegraphics[width=0.24\linewidth]{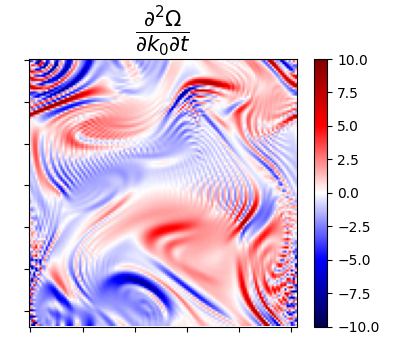}
    \includegraphics[width=0.24\linewidth]{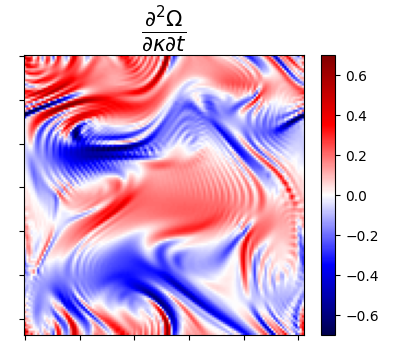}
    \includegraphics[width=0.24\linewidth]{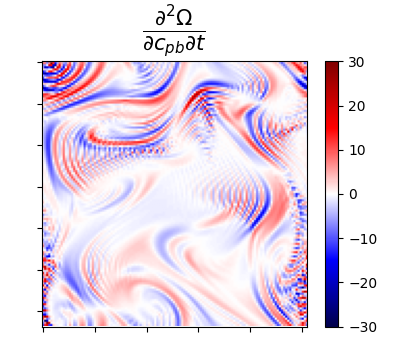}
    \caption{Effect of parameter variations on plasma dynamics on the vorticity $\Omega$. 
    From left to right, the varied parameters are $c_1$, $k_0$, $\kappa$, and $c_{pb}$. 
    Variations in different parameters lead to qualitatively distinct dynamical behaviors.
    This motivates the inclusion of all parameters in the FI-Conv experiments.
    }
    \label{fig:parameterEffect}
\end{figure}

Most existing studies (e.g., Refs.~\cite{Gahr2024HWROM,Artigues2025HW2D}) focus on the evolution of the plasma state by varying the parameter $c_1$, while typically keeping $\kappa$ and $c_{pb}$ fixed at unity. 
However, as illustrated in Fig.~\ref{fig:parameterEffect}, we observe that varying $\kappa$ and $c_{pb}$ as well influence the plasma dynamics in a manner that is qualitatively distinct from the effects of $c_1$.
Accordingly, our FI-Conv experiments in this paper are designed to incorporate variations across $c_1$, $k_0$, $\kappa$, and $c_{pb}$, allowing the model to capture a more diverse range of behaviors.


We generate a database of 320 simulations using the HW2D code by varying four parameters: $c_1$, $k_0$, $\kappa$, and $c_{pb}$. 
To the best of our knowledge, the HW system has not previously been studied under simultaneous multivariate parametric variations. 
Consequently, we restrict the present analysis to relatively small parameter intervals to assess the feasibility and potential of the proposed FI-Conv framework.
Each parameter is sampled independently from a uniform distribution over the following ranges: $c_1 \in [0.9, 1.1]$, $k_0 \in [0.55, 0.65]$, $\kappa \in [0.9, 1.1]$, and $c_{pb} \in [0.9, 1.0]$. 
Broader parameter ranges are reserved for future investigation.
Larger parameter variations can lead to distinct plasma geometry profiles during the turbulent phase and may require different numbers of time steps to reach turbulence. 
Accurately capturing these effects would necessitate additional simulations to appropriately select the domain size parameter $k_0$ and the numerical time step.

The spatial domain is discretized using a uniform $128\times128$ Cartesian grid, with a time step of $0.005$ and a total of 40,000 steps per simulation. 
Each simulation instance requires approximately 1,200 seconds on a single core of an Intel Xeon 6248R CPU, which is mounted on the Grace supercomputer in TAMU HPRC \footnote{https://hprc.tamu.edu/kb/User-Guides/Grace/}. 


Among the 320 simulation instances, 240 are used for training and 80 for testing. 
For each of the training simulation instances, the first 20,000 time steps are discarded, as they include the initial and transient growth phases before the plasma reaches the quasi-stable, saturated turbulent regime.
To generate training data, the remaining time steps of the simulation trajectories are separated into snapshot pairs. 
For each pair, the first snapshot serves as the neural network input, while the second snapshot is the prediction target. 
The time difference between snapshots is limited to at most one unit of time to ensure accurate learning of short-term dynamics. 
Longer-term predictions are obtained by iteratively applying the network in an autoregressive fashion, using each predicted state as input for the next step. 

\section{FI-Conv training}\label{sec:hyperparameter}

In our experiments with the HW model, the input to FI-Conv consists of 8 channels: the first three channels correspond to the plasma state variables $\Omega$, $\phi$, and $n$; the fourth channel encodes the time step between the input and target states for prediction $\Delta t_{\mathrm{i}}$; and channels five through eight contain the PDE parameters $(c_1, k_0, \kappa, c_{pb})$. 
The network output comprises two channels, representing the predicted plasma state variables $\Omega$ and $n$.

The training loss for FI-Conv is defined as the MSE:
\begin{equation}\label{eq:mseloss}
    \mathrm{MSE} = \frac{1}{100 N}\sum_{i=1}^N (\Omega_{\mathrm{pred},i} - \Omega_{\mathrm{true},i})^2 + \frac{1}{20 N}\sum_{i=1}^N (n_{\mathrm{pred},i} - n_{\mathrm{true},i})^2 \,,
\end{equation}
where $N$ denotes the batch size. 
The normalization coefficients of 100 for $\Omega$ and 20 for $n$ are chosen to mitigate order-of-magnitude differences between the variables during training.  
The loss is minimized using the AdamW optimization algorithm~\cite{Loshchilov2017AdamW} with a constant learning rate of 3$\times 10^{-4}$. 
We train the FI-Conv with the learning rates of $3\times 10^{-3}$, $1\times 10^{-3}$, $3\times 10^{-4}$, or $1\times 10^{-4}$, and found the learning rate $3\times 10^{-4}$ shows the lowest MSE value on the testing data set. 
Training FI-Conv for 14 epochs required approximately 80 hours on a single NVIDIA A100 GPU mounted on Grace supercomputer in TAMU HPRC \footnote{https://hprc.tamu.edu/kb/User-Guides/Grace/}, the batch size is set to 30 to accommodate the 40~GB GPU memory space. 

Following~\cite{Hao2024HardPINN}, we enforce a hard initial-condition constraint on the FI-Conv output:
\begin{eqnarray}
    \Omega_{\mathrm{pred}}(\Delta t_{\mathrm{i}}) &=& \Delta t_{\mathrm{i}} \, \Omega_{\mathrm{raw}} \times 100 + \Omega_{\mathrm{in}} \,, \\ 
    n_{\mathrm{pred}}(\Delta t_{\mathrm{i}}) &=& \Delta t_{\mathrm{i}} \, n_{\mathrm{raw}} \times 20 + n_{\mathrm{in}} \,,
\end{eqnarray}
where $\Omega_{\mathrm{raw}}$ and $n_{\mathrm{raw}}$ correspond to the first and second channels of the FI-Conv output, respectively, and $\Omega_{\mathrm{in}}$ and $n_{\mathrm{in}}$ denote the vorticity and density of the input plasma field. 
This process is highlighted in purple in Fig.~\ref{fig:FI-Conv-Detail}.


Due to the large state dimension of the discretized HW model and the small numerical time step, the full training dataset occupies approximately 700~GB of storage. 
As a result, each training epoch of FI-Conv requires roughly six hours. 
The total computational cost of generating the HW simulation dataset is approximately 100 CPU core-hours, which is lower than the cost of training FI-Conv.
However, for future applications of FI-Conv to more complex PDE systems (e.g., JOREK~\cite{Hoelzl2021Jorek}), the computational requirements for numerical simulations may be substantially higher, therefore limiting the size of the available training data.

\begin{table}[]
    \centering
    \begin{tabular}{c|c|c|c|c}
         Approach & Parameters & MSE & Inference Time (s) & Batch Time (s)  \\
         \hline
         FI-Conv                          & 31,004,978 & $6.73\times 10^{-6}$ & 0.047 & 0.224 \\
         U-Net \cite{Ronneberger2015Unet} & 42,354,386 & $7.52\times 10^{-5}$ & 0.064 & 0.094 \\
         FNO-16  \cite{Li2020FNO}            &  9,590,082 & $2.99\times 10^{-4}$ & 0.038 & 0.214 \\
         FNO-8  \cite{Li2020FNO}             &  4,804,162 & $3.21\times 10^{-4}$ & 0.024 & 0.141 \\
    \end{tabular}
    \caption{The comparison of the neural network number of parameters, MSE on the testing data set measured using Equation \ref{eq:mseloss}, inference time per instance, and training time per batch. The inference time and the training time are measured on a NVIDIA A100 GPU. Each of the neural network is trained on a NVIDIA A100 GPU for 80 hours. }
    \label{tab:nnTable}
\end{table}

Before proceeding, we compare FI-Conv, U-Net, and FNO in terms of MSE for forward prediction on the HW test dataset described in Section~\ref{sec:HW_equations}.
Table~\ref{tab:nnTable} reports the number of trainable parameters, MSE on the HW test dataset, inference time per instance, and training time per batch for FI-Conv, U-Net, and FNO. 
Among the three methods, FI-Conv achieves the lowest MSE. 
While FNO provides the fastest inference time, its prediction accuracy is inferior to that of FI-Conv. 
Increasing the number of retained Fourier modes in FNO from 16 to 32 may improve accuracy; however, this significantly increases GPU memory requirements during training and exceeds the capacity of a single NVIDIA A100 GPU with 40~GB of memory.

To obtain a more comprehensive assessment of FI-Conv, we will also study its sensitivity to training dataset size and emulate realistic data-limited scenarios using two reduced-data configurations.
The first, termed \emph{Reduced Instances}, decreases the number of training trajectories from 240 to 80. 
The second, referred to as \emph{Reduced Sampling}, retains all 240 training trajectories but randomly selects 6{,}000 snapshots per trajectory (corresponding to 30\% of the available snapshots) for training. 
Both configurations result in training datasets that are approximately 33\% of the size of the baseline training dataset with 240 trajectories with 20{,}000 snapshots. 
In both reduced-data configurations, the same set of 80 trajectories from the original setup is used for testing.

Table~\ref{tab:reducetrain} reports the MSE on the testing dataset for FI-Conv forward predictions under the different training scenarios. 
As expected, reducing the number of training instances leads to an increase in MSE, whereas reducing the number of snapshots per instance has only a marginal impact on predictive accuracy.
Overall, FI-Conv maintains strong performance even with reduced training data, achieving a minimum MSE of $1.77 \times 10^{-5}$.
In the following, we will present in more detail our results for forward prediction with FI-Conv, trained on 240 trajectories containing 20{,}000 snapshots and tested on 80 trajectories.

\begin{table}[]
    \centering
    \begin{tabular}{c|c|c|c}
         Model Name        & No.~Training Instances & Sampled Data & MSE \\
         \hline
         FI-Conv           & 240 & 20000 & $6.73\times 10^{-6}$ \\
         Reduced Instances & 80  & 20000 & $1.77\times 10^{-5}$ \\
         Reduced Sampling  & 240 & 6000  & $7.34\times 10^{-6}$
    \end{tabular}
    \caption{The MSE value in the testing data set of the FI-Conv trained on the original training data set, and the two reduced training data sets. The first column is the name of the models, the second column is the number of numerical simulation instances used in the training data set, the third column is the number of plasma snapshots sampled from each instance in the training data set, the fourth column is the MSE measured on the testing data set. }
    \label{tab:reducetrain}
\end{table}

\section{ FI-Conv for forward prediction}\label{sec:forward}

This section reports our numerical results focused on forward prediction of plasma state evolution using the proposed FI-Conv model.
The subsequent section addresses the inverse problem of parameter estimation from sample data, using the same trained model without further weight updates.

\begin{figure}[t!]
    \centering
    \includegraphics[width=\linewidth]{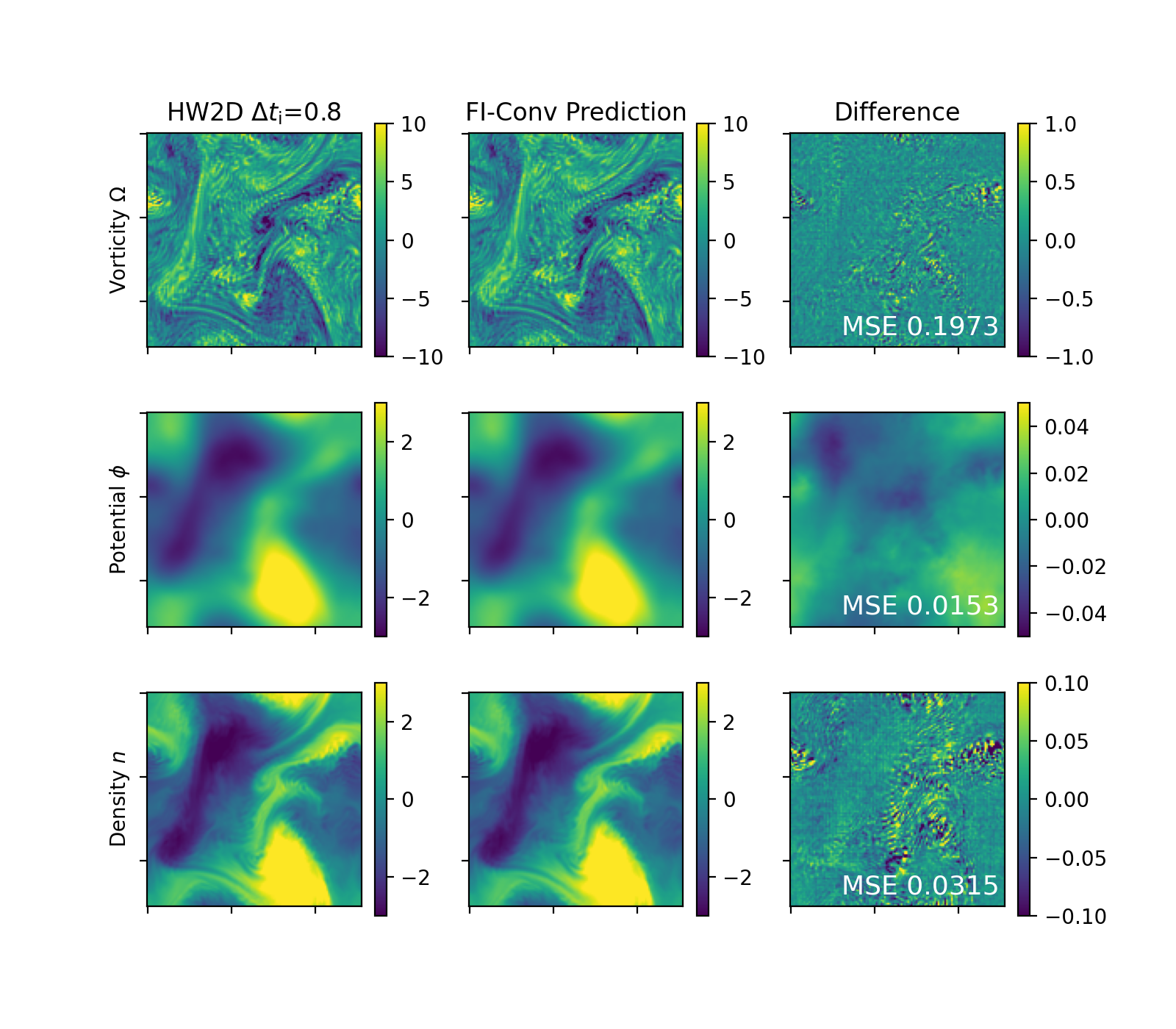}
    \caption{One-step prediction of the plasma state at 0.8 time units after an input initial state in the testing data. Left column: reference HW2D simulation code results. Middle column: FI-Conv prediction. Right column: difference between the HW2D results and the FI-Conv prediction. The time after the FI-Conv input state is $\Delta t_{\mathrm{i}}=0.8$. }
    \label{fig:oneSnap}
\end{figure}

Starting from a given initial plasma state, the trained FI-Conv model performs one-step forward prediction of the plasma state at arbitrary evolution times $\Delta t_{\mathrm{i}} \in [0,1]$. 
Figure~\ref{fig:oneSnap} compares the predicted vorticity $\Omega$, electrostatic potential $\phi$, and density $n$ fields with the corresponding reference solutions from the HW2D code at $\Delta t_{\mathrm{i}} = 0.8$, for one trajectory in the testing set.

To predict plasma evolution over longer time horizons, i.e., with $\Delta t_{\mathrm{i}} > 1$ we adopt an autoregressive strategy~\cite{Gopakumar2024FNOPlasma}. 
A fixed time increment $t_{\mathrm{a}} \in [0,1]$ is specified, such that FI-Conv advances the plasma state by $t_{\mathrm{a}}$ at each step. 
The first prediction is obtained by applying FI-Conv to the initial plasma state, while subsequent predictions use the previously predicted state as input. 
This process is repeated until the desired final evolution time is reached. 

\begin{figure}[t!]
    \centering
    \includegraphics[width=\linewidth]{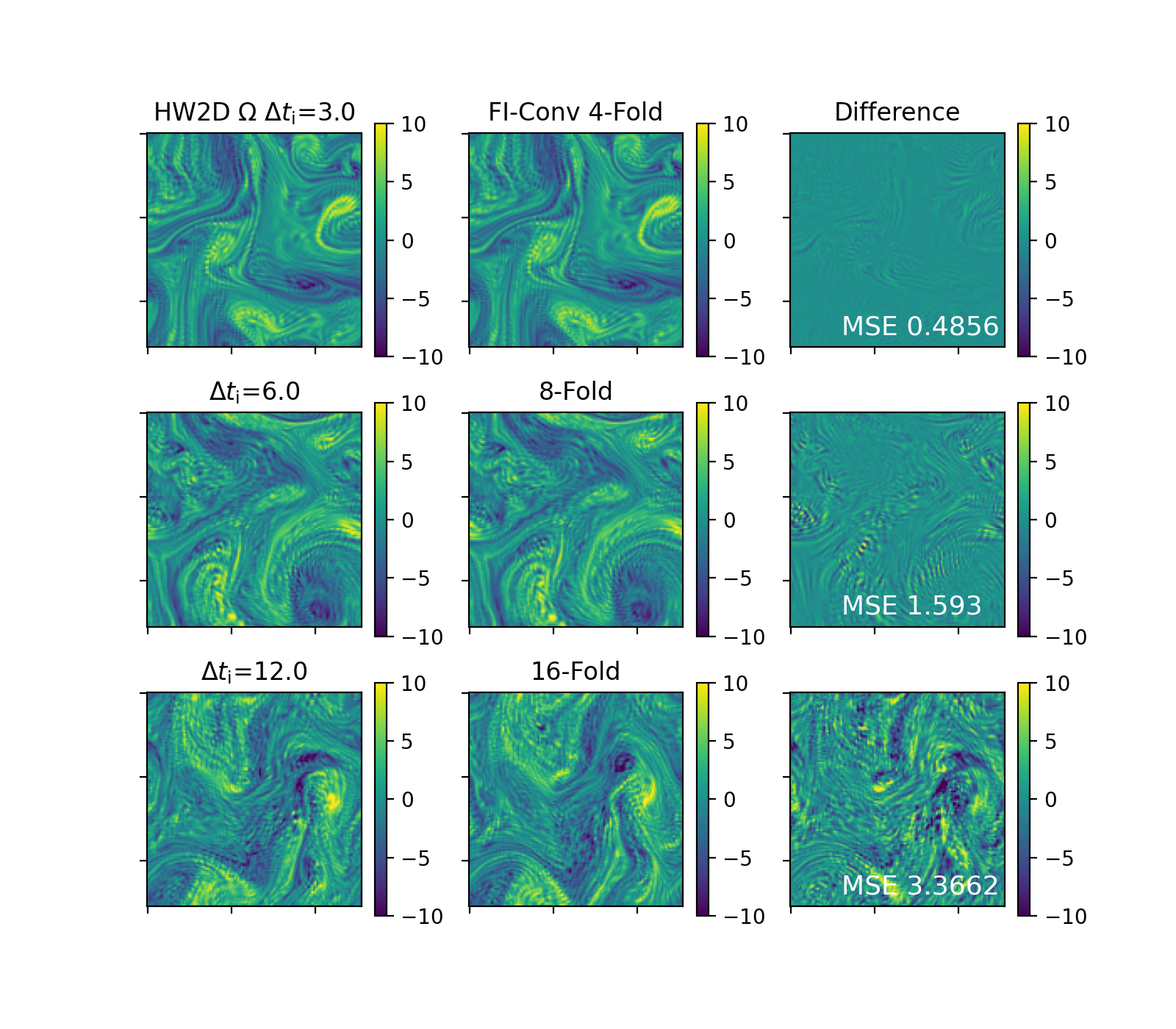}
    \caption{Autoregressive prediction of plasma state with time step $t_{\mathrm{a}} = 0.75$ from an input initial state in the testing data. Left column: reference $\Omega$ snapshots computed with the HW2D simulation code. Middle column: $\Omega$ snapshots predicted by FI-Conv after 4, 8, and 16 autoregressive steps. Right column: difference between the HW2D results and the FI-Conv prediction. From upper to lower, the time after the FI-Conv input state $\Delta t_{\mathrm{i}}$ is 3.0, 6.0, and 12.0, respectively. }
    \label{fig:multiSnap}
\end{figure}

An example of an autoregressive prediction for the vorticity field $\Omega$ is shown in Fig.~\ref{fig:multiSnap}. 
The FI-Conv predictions begin to deviate from the high-fidelity HW2D solution at approximately $\Delta t_{\mathrm{i}} = 6$, corresponding to eight autoregressive steps with $t_{\mathrm{a}}=0.75$. 
We note that such divergence at long integration times is not indicative of an algorithmic failure. 
In chaotic systems such as drift-wave turbulence governed by the HW model, even infinitesimal perturbations in the initial state can grow exponentially, leading to de-correlation between two otherwise valid simulations over sufficiently long horizons.
Consequently, pointwise agreement with a reference solution becomes less meaningful at long times, and more meaningful comparisons are based on derived quantities that capture the statistical properties of the turbulence, which we examine next.

In the next experiment, FI-Conv is trained using HW2D simulation data spanning $t \in [100, 200]$, during which the plasma fields transition from the initial transient growth phase to the development of turbulent structures. 
Prior studies~\cite{Gahr2024HWROM} reported that the plasma dynamics may not reach a statistically stationary turbulent regime until approximately $t = 300$. 
To evaluate whether FI-Conv can capture physically meaningful quantities over extended evolution times, we perform a reference HW2D simulation with parameters $c_1 = 1$, $k_0 = 0.6$, $\kappa = 1$, and $c_{pb} = 1$ up to $t = 600$ with a numerical time step of 0.003. 
FI-Conv is then used to extrapolate the plasma evolution starting from $t = 300$ (i.e., after the onset of turbulence), with an autoregressive time step of $t_{\mathrm{a}} = 1$.

\begin{figure}
    \centering
    \includegraphics[width=\linewidth]{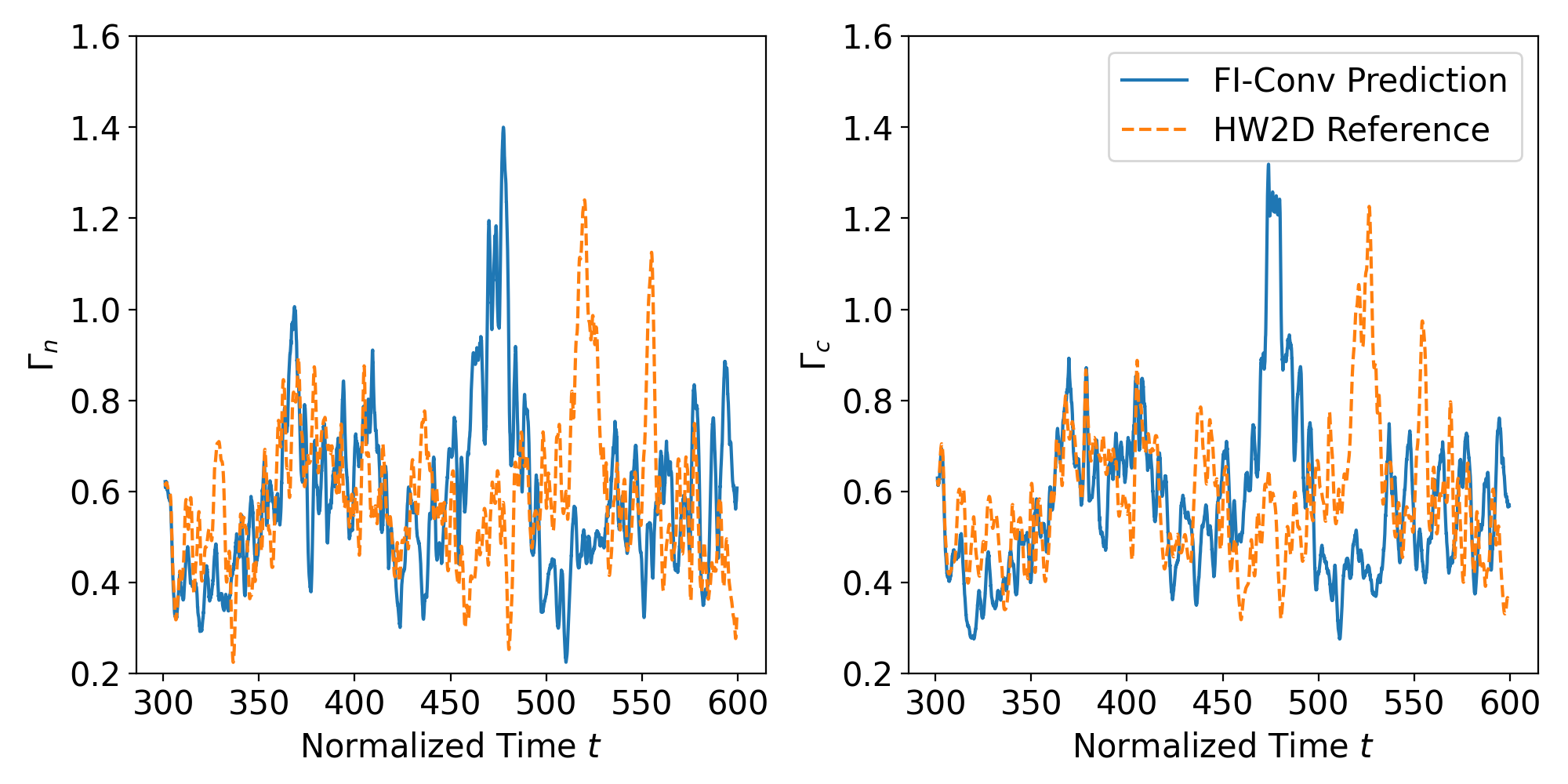}
    \caption{The evolution of $\Gamma_c$ and $\Gamma_n$ derived quantities of a numerical simulation instance starting at $t=300$ and ending at $t=600$. 
    The PDE parameters are $c_1=1$, $k_0=0.6$, $\kappa=1$, $c_{pb}=1$. }
    \label{fig:physQuan}
\end{figure}

Figure~\ref{fig:physQuan} shows the time evolution of the particle and heat fluxes, $\Gamma_n$ and $\Gamma_c$, predicted by FI-Conv and obtained from the high-fidelity HW2D simulation over $t \in [300, 600]$, i.e., in the temporal extrapolation regime. 
The corresponding MSEs, computed over successive time intervals, are reported in Table~\ref{tab:gammaNCInterval}. 
The predictions closely match the numerical results at early times, with MSE values of approximately 0.07 over the first $\sim$15 time units. 
As expected for long autoregressive rollouts in a turbulent regime, the error gradually increases with integration time, reaching $\sim$0.29 in the intervals $t \in [400, 500]$ and $t \in [500, 600]$. 
Despite this growth in error, the predicted fluxes preserve the overall temporal trends and fluctuation levels of the reference solution.

\begin{figure}
    \centering
    \includegraphics[width=\linewidth]{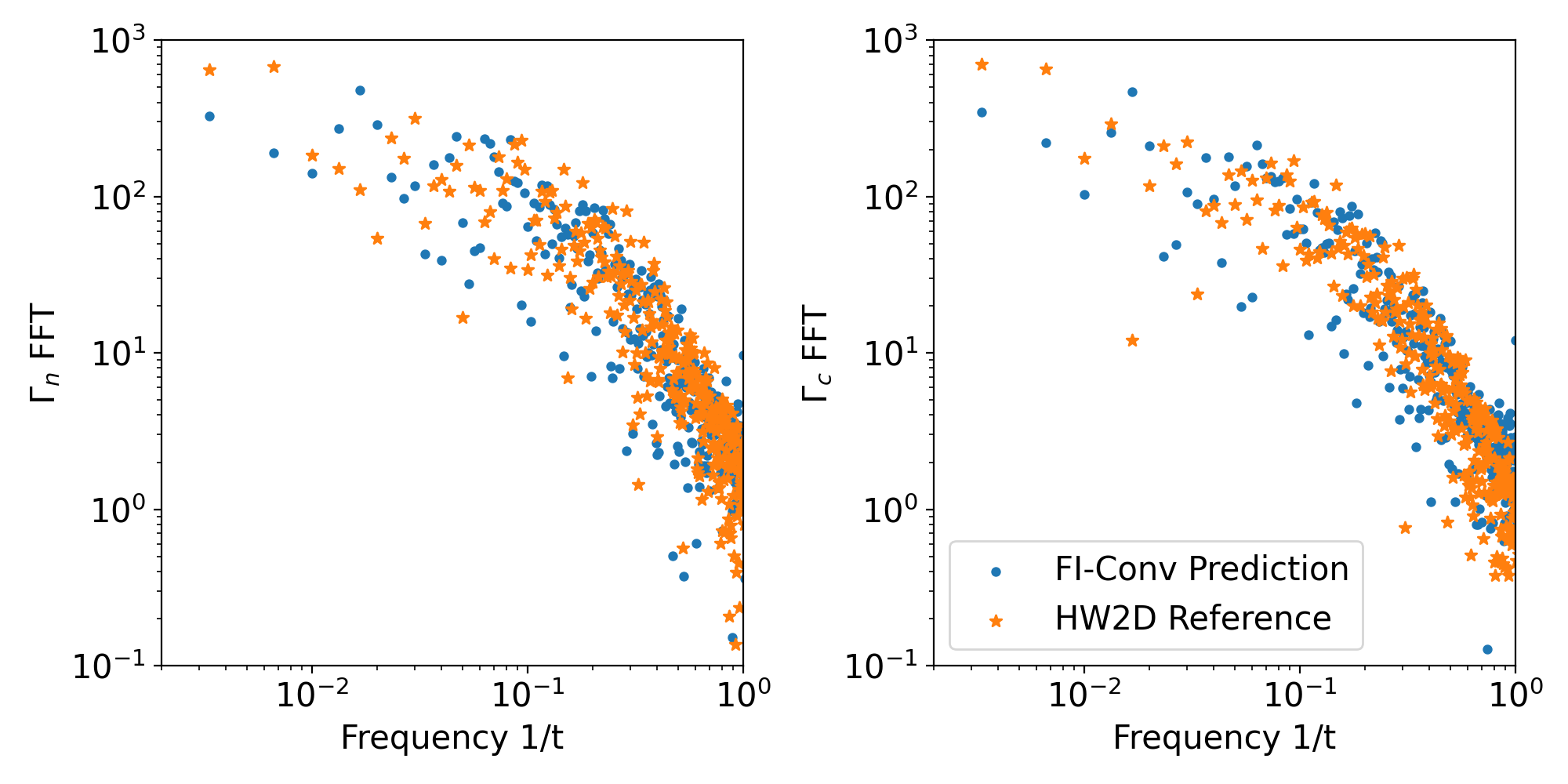}
    \caption{Fourier transform magnitudes of the $\Gamma_c$ and $\Gamma_n$ physical parameters between $t=300$ and $t=600$, using the data in Figure \ref{fig:physQuan}.}
    \label{fig:physQuanFFT}
\end{figure}

To further assess the long-term prediction accuracy and stability of FI-Conv, we compute the temporal mean and standard deviation of $\Gamma_n$ and $\Gamma_c$, reported in Table~\ref{tab:gammaNCMean}. 
Additionally, Fig.~\ref{fig:physQuanFFT} shows the Fourier transform magnitudes of $\Gamma_n$ and $\Gamma_c$ over the interval $t \in [300, 600]$, allowing us to evaluate whether FI-Conv captures the oscillatory behavior of the turbulent fluxes. 
The temporal means and standard deviations predicted by FI-Conv closely match the reference values, and the Fourier magnitudes of both fluxes are in good agreement with the reference numerical results. 
These comparisons indicate that FI-Conv captures the statistical and spectral characteristics of the plasma turbulence with good accuracy, demonstrating stable long-time predictions despite the chaotic dynamics of the HW system.

\begin{table}[htb]
    \centering
    \begin{tabular}{c|c|c}
        Time Intervals & MSE for $\Gamma_n$ & MSE for $\Gamma_c$ \\ 
        \hline
        $300\leq t\leq315$    & 0.0709             & 0.0693             \\
        $315\leq t\leq350$    & 0.1533             & 0.1237             \\
        $350\leq t\leq400$    & 0.1330             & 0.0846             \\
        $400\leq t\leq500$    & 0.2913             & 0.2761             \\
        $500\leq t\leq600$    & 0.2907             & 0.2783             \\
    \end{tabular}
    \caption{\label{tab:gammaNCInterval} The average and the standard deviation of $\Gamma_n$ and $\Gamma_c$ values between $t=300$ and $t=600$, from the reference numerical solver and the FI-Conv predicted plasma field, as shown in Figure \ref{fig:physQuan}. }
\end{table}

\begin{table}[htb]
    \centering
    \begin{tabular}{c|c|c|c}
                                    &                     & Reference & FI-Conv  \\
        \hline
        \multirow{2}{*}{$\Gamma_n$} & Mean                & 0.584     & 0.585 \\
                                    & Standard Deviation  & 0.161     & 0.187 \\
        \hline
        \multirow{2}{*}{$\Gamma_c$} & Mean                & 0.585     & 0.573 \\
                                    & Standard Deviation  & 0.147     & 0.176   
    \end{tabular}
    \caption{\label{tab:gammaNCMean} The average and the standard deviation of $\Gamma_n$ and $\Gamma_c$ values between $t=300$ and $t=600$, from the reference HW2D numerical solver and the FI-Conv predicted plasma field, as shown in Figure \ref{fig:physQuan}. }
\end{table}

\begin{figure}
    \centering
    \includegraphics[width=0.3\linewidth]{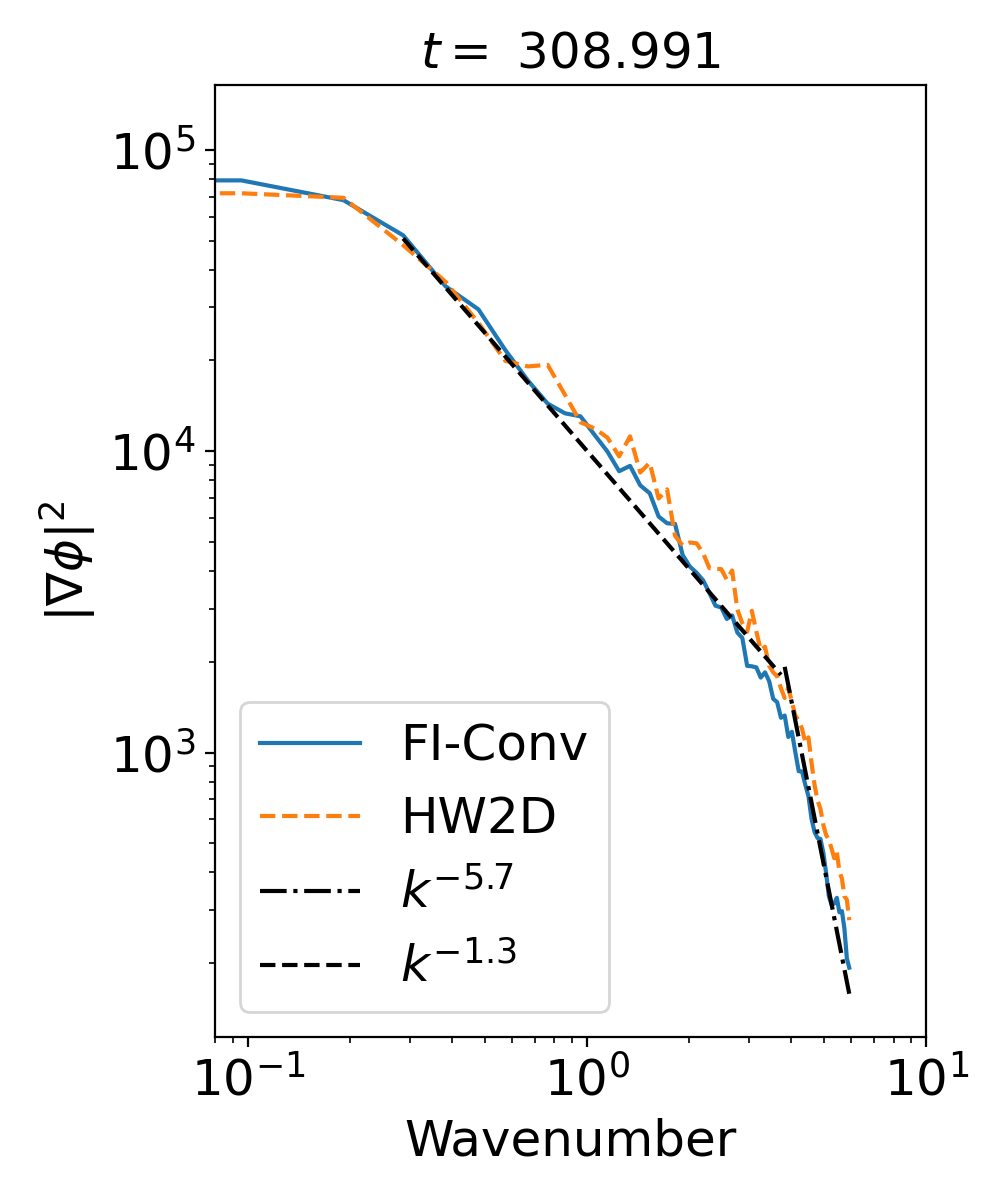}
    \includegraphics[width=0.3\linewidth]{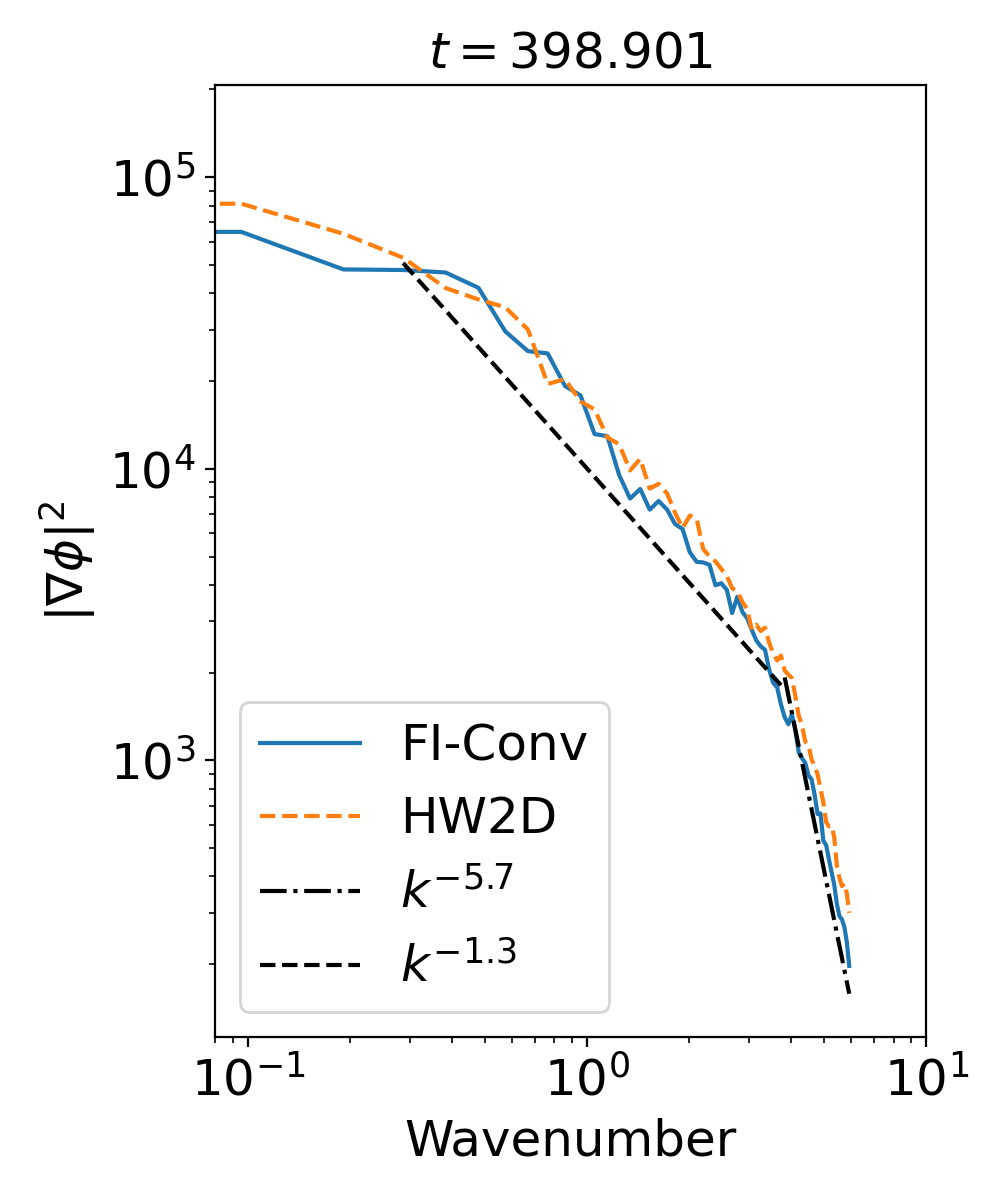}
    \includegraphics[width=0.3\linewidth]{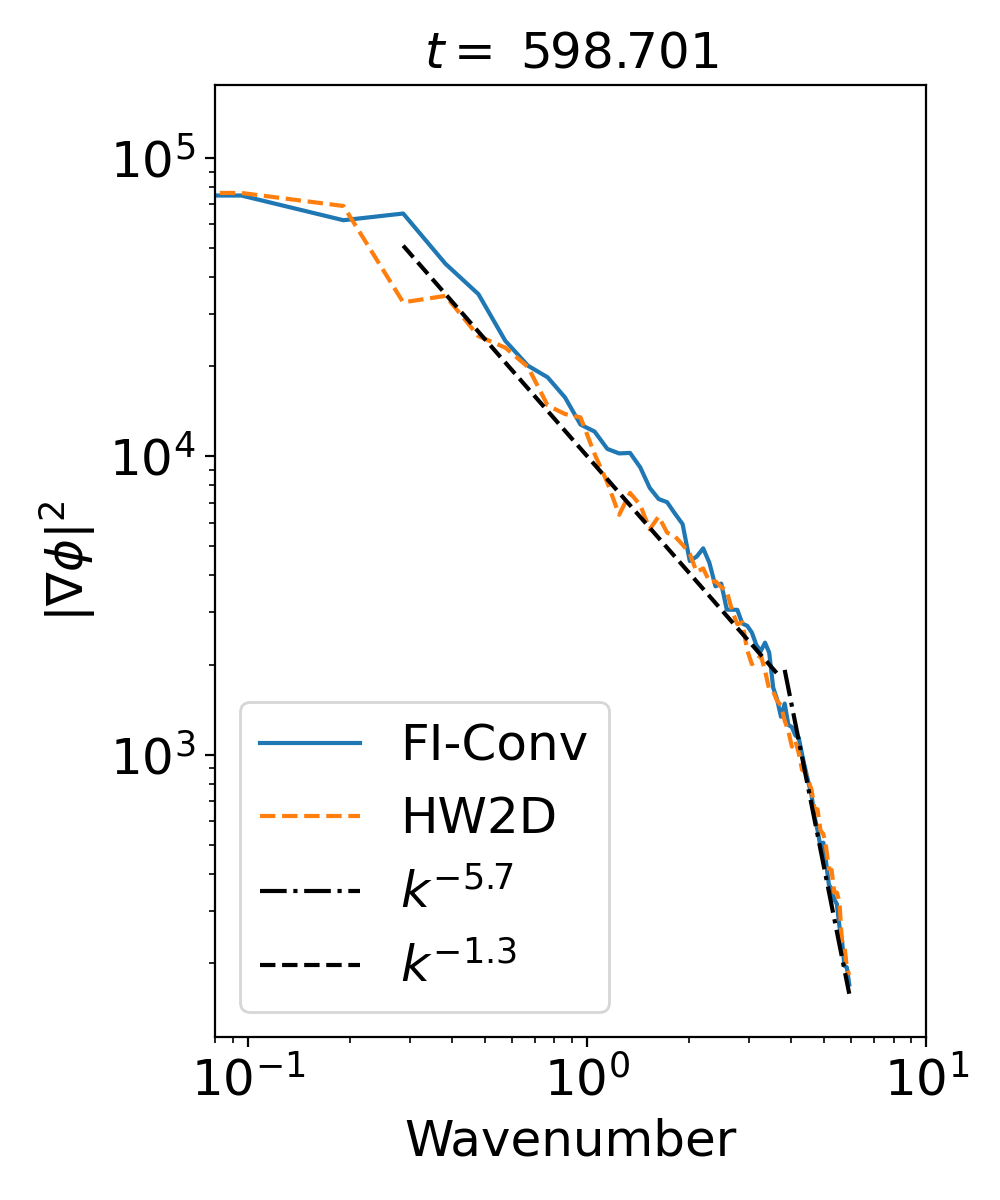}
    \caption{Spatial Fourier transform magnitudes of $|\nabla\phi|^2$ at different times in the evolution. The black dashed dash-dotted lines display the power law relations $k^{-1.3}$ and $k^{-5.7}$ of the wavenumber.}
    \label{fig:spatialFFT}
\end{figure}

Using the FI-Conv predicted plasma fields over $t \in [300, 600]$, we compute the Fourier spectrum of $|\nabla \phi|^2$ \cite{Majda2018SpaFour} as a metric for assessing prediction accuracy over long integration times. 
The results are shown in Fig.~\ref{fig:spatialFFT}. 
The FI-Conv predictions reproduce the decay of Fourier amplitudes observed in the numerical simulations up to $t = 400$, with only slight deviations at the final simulation time $t = 600$. 
Additional simulations with the same PDE parameters but different initial conditions, presented in \ref{sec:app:moreQuan}, demonstrate similar prediction accuracy for $\Gamma_n$ and $\Gamma_c$, confirming the robustness of FI-Conv across multiple initial condition realizations. 
Importantly, the FI-Conv predictions maintain consistency with the numerical results even at high spatial frequencies. 
This preservation of fine-scale structures is attributed to the skip-connection architecture in the U-Net, which transmits high-resolution information across layers. 
This is in contrast to previous machine-learning approaches applied to the HW system, which did not achieve similar consistency, due either to reduced-order modeling \cite{Garrido2025Reduce} or down-sampled training data \cite{Artigues2025HW2D}.

\section{ FI-Conv for the Inverse Problem}\label{sec:inverse}

This section presents results for solving inverse problems using the proposed FI-Conv framework.
Specifically, we demonstrate that a trained FI-Conv can estimate the HW parameters $c_1$, $k_0$, $\kappa$, and $c_{pb}$ from a small set of plasma snapshots, without requiring any updates to the network weights.

The inverse parameter estimation is carried out as follows.
Plasma snapshots are paired into input and output states separated by a prescribed time interval.
The PDE parameters $c_1$, $k_0$, $\kappa$, and $c_{pb}$ are initialized with some initial guesses.
FI-Conv then predicts the output snapshots using the input snapshots and the current parameter values, and the MSE between the predictions and the reference outputs is computed via Eq.~\ref{eq:mseloss}.
Gradients of the MSE with respect to the PDE parameters are used to update these parameters so as to minimize the loss.
Throughout this process, only the PDE parameters are updated, while the network weights remain fixed.
Parameter updates are performed using the AdamW optimizer~\cite{Loshchilov2017AdamW}. 

\begin{figure}
    \centering
    \includegraphics[width=\linewidth]{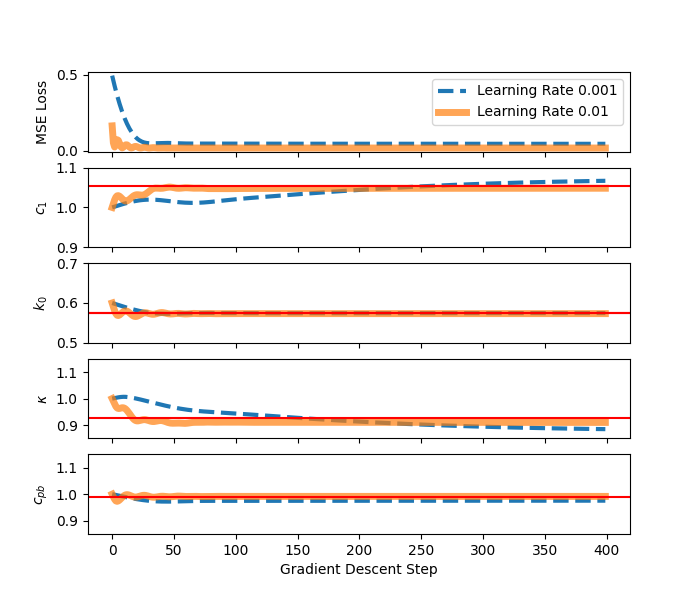}
    \caption{An example of solving the inverse problem. 
    The red horizontal line marks the truth of the PDE parameters $c_1$, $k_0$, $\kappa$, $c_{pb}$. 
    The blue line in the first panel shows the MSE value at each gradient descent step, while the blue lines in the lower four panels display the evolution of the predicted PDE parameters. }
    \label{fig:inverse}
\end{figure}

Figure~\ref{fig:inverse} illustrates an example of the inverse problem solved using 32 randomly selected pairs of plasma snapshots, comparing results obtained with learning rates of 0.01 (orange line) and 0.001 (blue dashed line).
A total of 400 gradient descent steps are performed, with a total computation time of 202 seconds on an NVIDIA A100 GPU per solution. 
When a large learning rate (i.e., 0.01) is chosen, the predicted PDE parameters converge to the true values within approximately 100 steps, and the MSE loss decreases nearly monotonically throughout the optimization. 
When a large learning rate of 0.01 is used, the predicted PDE parameters converge to their true values within roughly 100 steps, and the MSE loss decreases nearly monotonically. 
In contrast, a smaller learning rate of 0.001 results in a higher final MSE and lower accuracy in parameter estimates. 
For this reason, we adopt a learning rate of 0.01 for solving the inverse problem across all test instances. 
Although this value is larger than typical neural network training rates, it helps the optimization avoid local minima, which also explains the small fluctuations observed in the MSE curve during the initial steps.

\begin{figure}
    \centering
    \includegraphics[width=\linewidth]{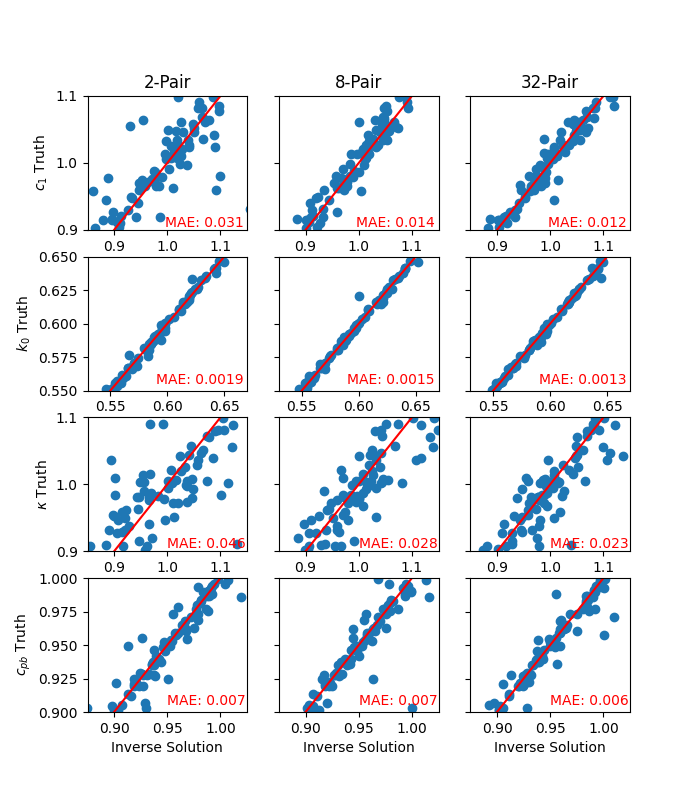}
    \caption{Predicted parameters versus true parameters using 80 simulation instances in the testing data set. The $c_{pred}=c_{true}$ line is displayed in red. The predicted parameters are on the x-axis, while the true parameters are on the y-axis. }
    \label{fig:invScatter}
\end{figure}

Figure~\ref{fig:invScatter} compares the true PDE parameters with the inverse solutions obtained for 80 test instances, using 2, 8, or 32 randomly selected pairs of plasma snapshots. 
To quantitatively assess the accuracy of the inverse solutions, we compute the mean absolute error (MAE):
\begin{equation}
    \mathrm{MAE} = \frac{1}{N} \sum_{i=1}^N \left| c_{\mathrm{true},i} - c_{\mathrm{pred},i} \right|\,,
\end{equation}
where $c_{\mathrm{true},i}$ and $c_{\mathrm{pred},i}$ denote the true and predicted values of the PDE parameters $c_1$, $c_{pb}$, $\kappa$, or $k_0$, respectively. 
The results indicate that $k_0$ and $c_{pb}$ can be estimated more accurately than $c_1$ and $\kappa$. 
Increasing the number of plasma snapshot pairs from 2 to 8 leads to a substantial improvement in parameter estimation accuracy, whereas further increasing from 8 to 32 pairs provides only marginal gains. 
This suggests that the FI-Conv inverse solution reaches a practical accuracy limit when 32 or more snapshot pairs are used.

\section{Conclusion}\label{sec:conclusion}

This paper presented a convolutional operator network architecture, FI-Conv, for both forward and inverse problems in time-evolving PDE systems with complex dynamics. 
FI-Conv enables long-term predictions using an autoregressive rollout approach, while remaining computationally efficient. 
Its performance was demonstrated on simulated data from a modified version of the Hasegawa--Wakatani equation system for plasma turbulence. 
The predicted plasma fields exhibit point-wise agreement with the HW2D numerical code up to approximately 10 time units. 
Beyond this horizon, the long-term evolution of derived physical quantities, $\Gamma_n$ and $\Gamma_c$, computed from the FI-Conv predictions, is consistent with the reference simulations in three aspects: (1) temporal mean, (2) temporal standard deviation, and (3) Fourier amplitude spectra.
We also proposed an inverse problem algorithm that accurately estimates the PDE parameters from time-shifted plasma snapshots without necessitating the modification of the network weights. 
The method constructs an MSE loss between predicted and reference snapshots and applies gradient descent on the input parameters using automatic differentiation. 
Across 80 testing instances, we observed that $k_0$ was the most challenging parameter to estimate, while $\kappa$ was the easiest. 
Accuracy improves with the number of snapshot pairs, reaching a practical limit at approximately 32 pairs.
The ability of FI-Conv to provide both forward predictions and inverse parameter estimation opens new avenues for plasma state diagnostics in tokamak devices \cite{Equipe1978Diagnostics}, as well as potential optimization of device configurations alongside reinforcement-learning methods \cite{Degrave2022TokamakRL}. 
Future work will explore the application of FI-Conv to full magneto-hydrodynamical (MHD) simulations, such as those generated by JOREK~\cite{Hoelzl2021Jorek}.

\section{Acknowledgement}
We gratefully acknowledge the high performance research computing resources provided by Texas A\&M University (https://hprc.tamu.edu). 
I.-G.~F.~was supported in part by the National Science Foundation under award DMS-2436357 and by the Academy of Data Science at Virginia Tech. 
X.~C.~is supported by the Texas A\&M University Institute of Data Science (TAMIDS). 
The work of U.~B-N.~is supported by the National Science Foundation under award CCF-2225507. 
D.H.~is supported by the Department of Energy under award DE-FG02-04ER54742. 
\appendix

\section{Further experiments for the quantities of interest}\label{sec:app:moreQuan}

\begin{figure}[htb!]
    \centering
    \includegraphics[width=\linewidth]{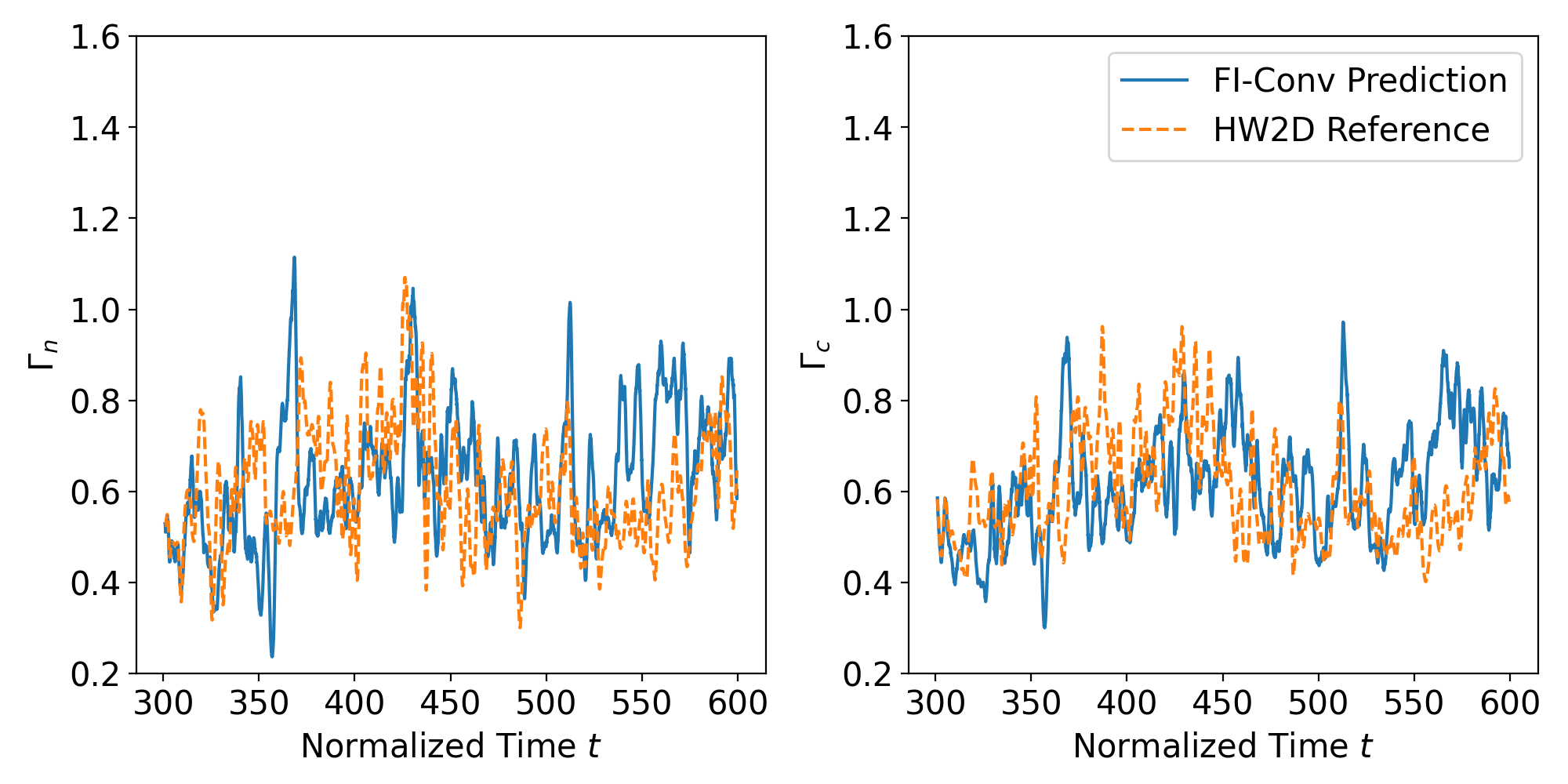}
    \includegraphics[width=\linewidth]{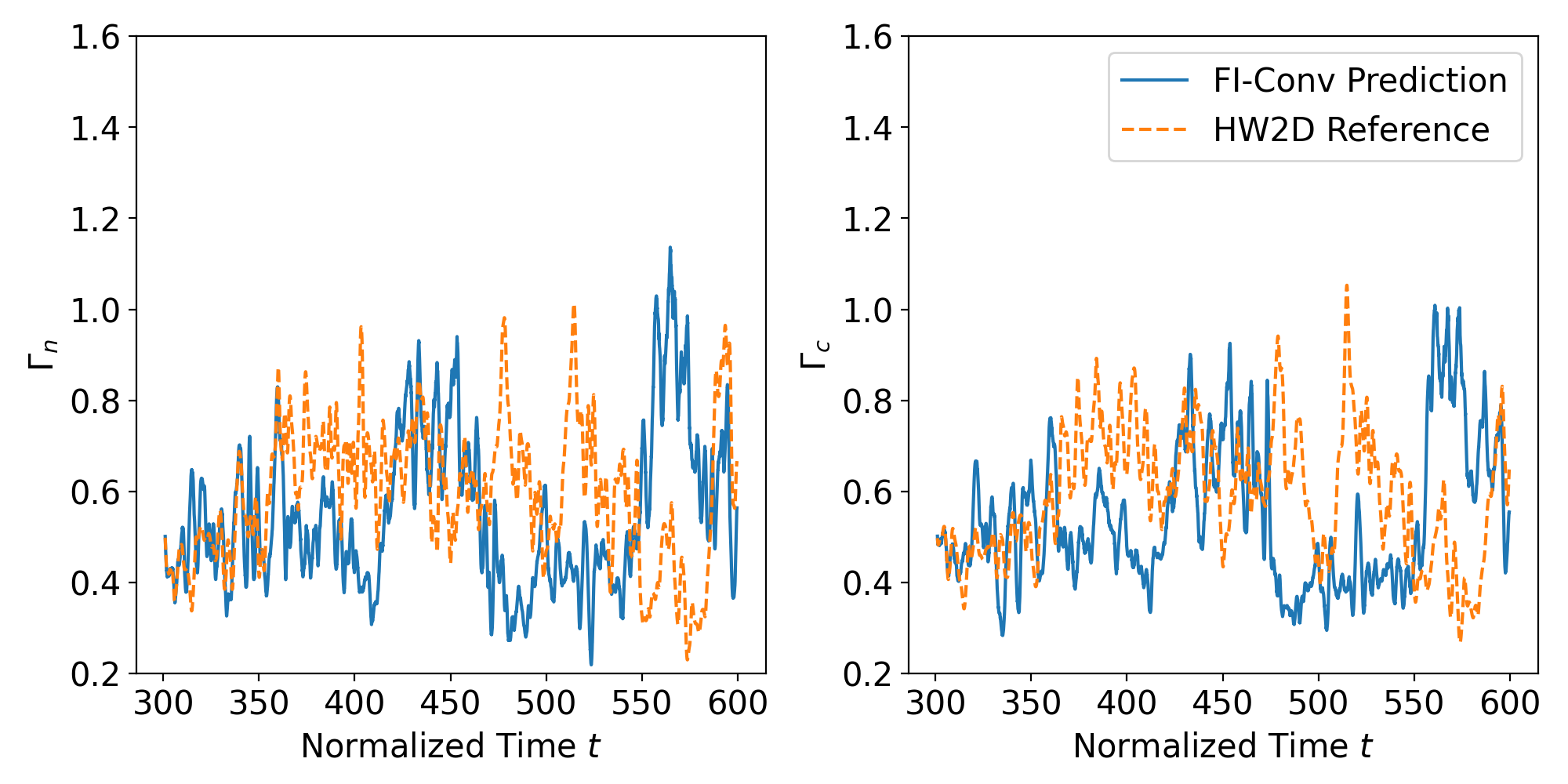}
    \caption{Same as Figure \ref{fig:physQuan}, but with different initial conditions. }
    \label{fig:PQExtra1}
\end{figure}

\begin{figure}
    \centering
    \includegraphics[width=\linewidth]{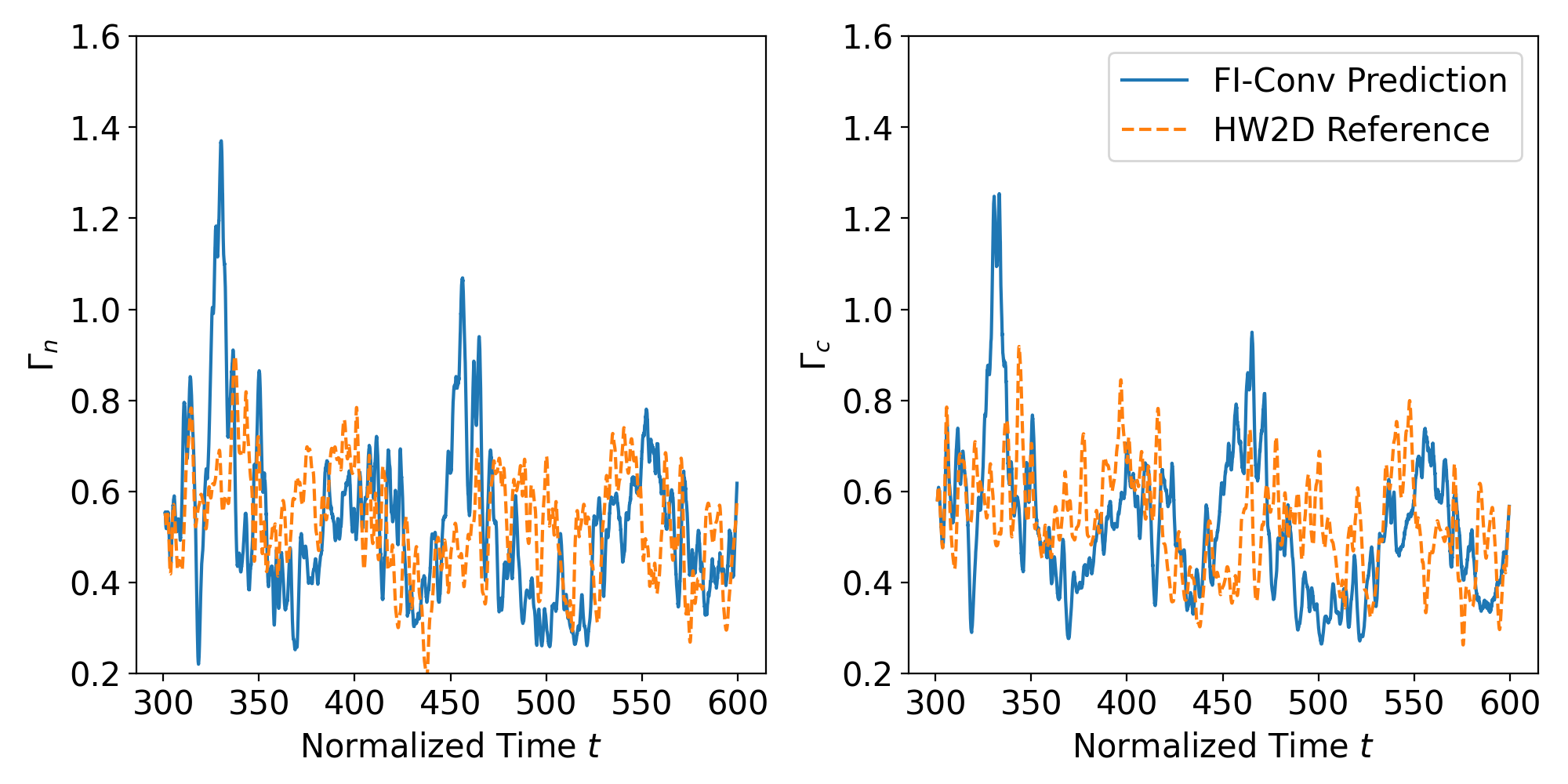}
    \includegraphics[width=\linewidth]{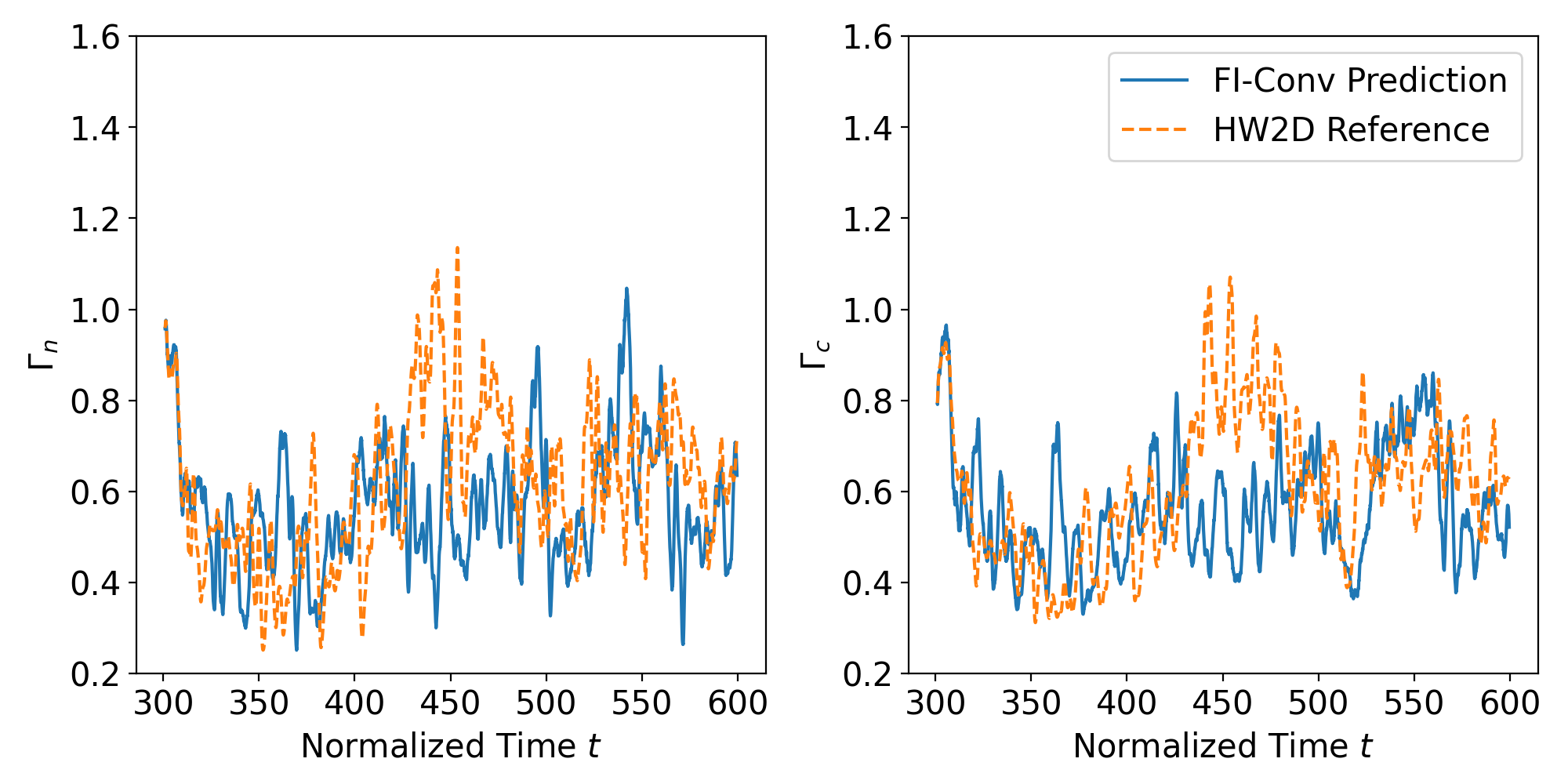}
    \includegraphics[width=\linewidth]{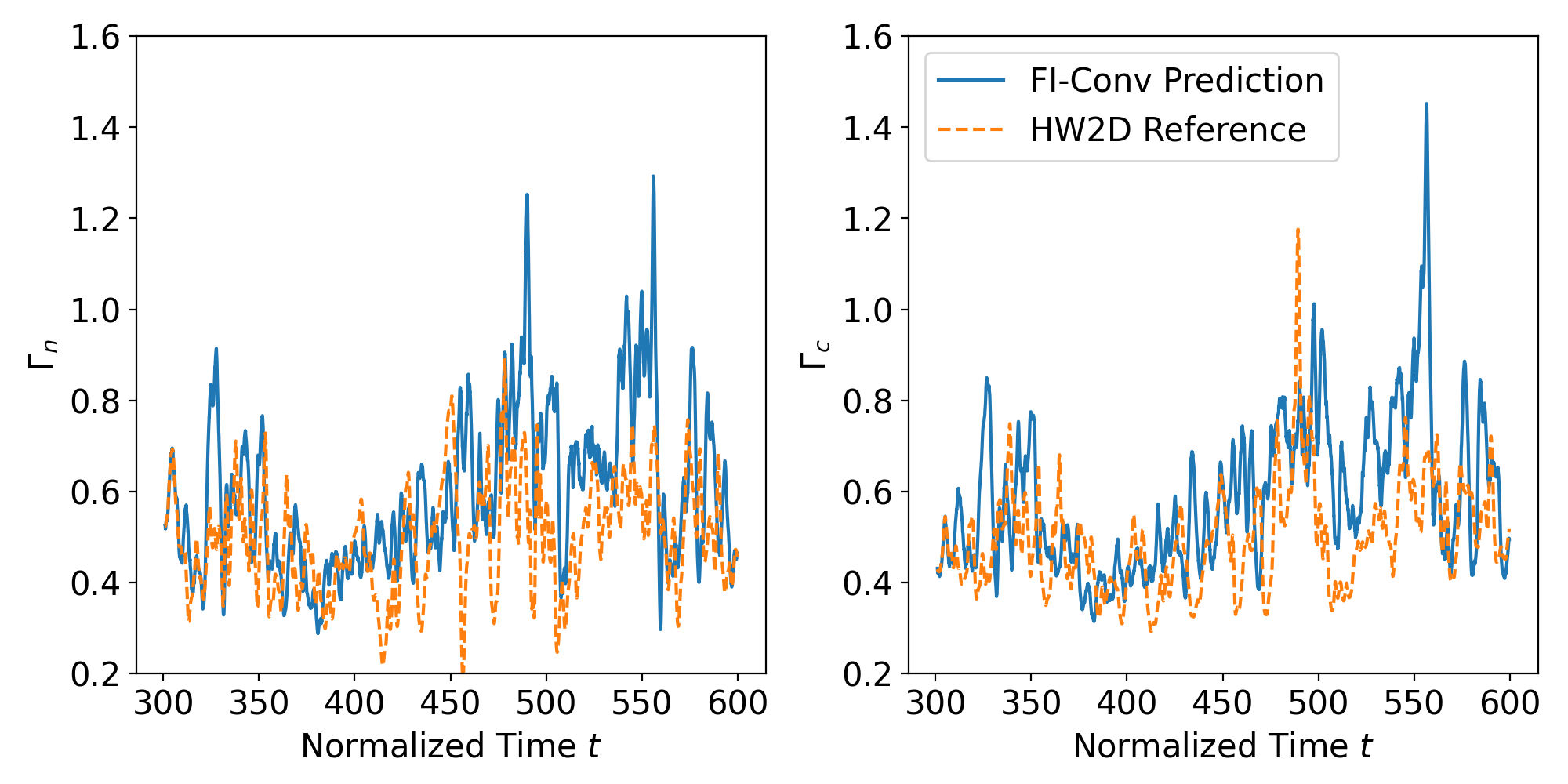}
    \caption{Same as Figure \ref{fig:physQuan}, but with different initial conditions. }
    \label{fig:PQExtra2}
\end{figure}

For a more detailed overview, we report further results as in Sec.~\ref{sec:forward} using different initial conditions. 
FI-Conv is used to predict the plasma state time evolution over $t \in [300, 600]$, allowing comparison of the derived quantities $\Gamma_n$ and $\Gamma_c$ with the reference results from the HW2D code. 
Figures~\ref{fig:PQExtra1} and \ref{fig:PQExtra2} show that FI-Conv maintains a similar level of accuracy across these additional simulation instances, consistent with the results reported in Sec.~\ref{sec:forward}.

\clearpage
\bibliographystyle{elsarticle-num} 
\bibliography{PIHDref}
\end{document}